# IBMMS Decision Support Tool For Management Of Bank Telemarketing Campaigns

Ali KELES and Ayturk KELES

Department of Computer Education and Instructional Technology, Faculty of Education
Agri Ibrahim Cecen University, TR-04100 Agri, Turkey
* Corresponding author: Ali KELES

## ABSTRACT

*Although direct marketing is a good method for banks to utilize in the face of global competition and the financial crisis, it has been shown to exhibit poor performance. However, there are some drawbacks to direct campaigns, such as those related to improving the negative attributes that customers ascribe to banks. To overcome these problems, attractive long-term deposit campaigns should be organized and managed more effectively.*

*The aim of this study is to develop an Intelligent Bank Market Management System (IBMMS) for bank managers who want to manage efficient marketing campaigns. IBMMS is the first system developed by combining the power of data mining with the capabilities of expert systems in this area. Moreover, IBMMS includes important features that enable it to be intelligent: a knowledge base, an inference engine and an advisor. Using this system, a manager can successfully direct marketing campaigns and follow the decision schemas of customers both as individuals and as a group; moreover, a manager can make decisions that lead to the desired response by customers.*

## 1. INTRODUCTION

The worldwide banking industry has experienced dramatic changes with respect to the way business is directed. The greater adoption of electronic banking, which allows for the capturing of transactional data, has become more practical, and the amount of such data has risen significantly. It is not possible for a human to analyse such a large amount of raw data and efficiently translate the data into useful knowledge for the benefit of an organization. The large amounts of data that banks have been collecting for years can have a significant influence on the success of Data Mining (DM) efforts. By using DM to analyse patterns and trends, bank managers can more precisely foresee customer reactions to modifications of interest rates, such as which customers will likely accept new product offers and which customers will have a higher risk of not paying a loan, thus making customer relationships more lucrative. Furthermore, banks can use DM to determine their most lucrative credit card customers or high-risk loan applicants [1].

In regard to communication, there are two opposing approaches in marketing: mass media marketing and direct marketing. In the case of mass media marketing, only one communication message is broadcast to all possible customers via various media, such as outdoor communication, magazines, newspapers, radio or television. This type of approach typically involves a significant amount of waste: only a small proportion of customers that are communicated with will show an interest in buying the product. When the competition increases and when more fragmentation occurs in the market, the importance of waste increases. In addition, even though there has been significant investment into market research and media planning, the value of mass media marketing remains difficult to ascertain. Under the most favourable conditions, indications about the number and type of people that are influenced by such marketing can be provided. However, data on customer response is typically not available.

The increased popularity of direct marketing has been a result of these developments, especially in the sectors of finance, banking, insurance, and telecommunication. Direct marketing ultimately aims to establish cost-effective, two-way and one-to-one communications with individual customers, which is not restricted to the internet; most direct marketing communication is still performed via traditional channels, such as direct mail, sms, email and inbound and outbound calls. To achieve efficient direct marketing, possessing information about the present and estimating future customer preferences is a fundamental requirement. In the current business climate, customer preferences change dynamically, and they are too complicated to derive direct conclusions [2].

Direct marketing campaigns are essential instruments for enhancing the economic gain of a firm in two respects: obtaining new customers and creating additional yield from present customers [3]. The availability of customer data combined with improvements in data analysis has prompted firms to develop more customer-oriented strategies in recent decades.

Different data-based approaches for direct marketing have been offered by both the marketing [4, 5, 6] and data-mining communities [7, 8].

Nie and others [9] applied two DM algorithms to develop a churn prediction model using credit card data: logistic regression and decision tree (DT). According to their test results, the performance of regression is slightly higher than that of a decision tree. The results indicate that the use of variables that do not have multi-colinearity lead to a better performance. However, models based on decision trees can provide easily understandable rules in rule form; the rules can provide guidance for banks in creating marketing strategies. Nie and others [9] believe that decision tree rules have indicated that customer behaviours can better predict future customer decisions.

Moro and others [10] used the CRISP-DM methodology for bank direct marketing campaigns. In the third iteration of CRISP-DM, they obtained good results using three DM algorithms (Nave Bayes, Decision Tree and Support Vector Machines).

DT is a famous method and has been successfully applied to numerous real-world problems [11]. It is a symbolic learning method whereby knowledge is obtained from a training dataset in a hierarchical form consisting of nodes and ramifications [12].

Numerous data analysis studies on marketing can be found in the literature. However, most of these studies do not go beyond performing analyses. Even in the best cases, it is not possible for these analyses to be used as supportive instruments in decision making tasks whereby managers develop their marketing strategies. One cannot expect managers to conduct analyses, which involve advanced technologies, and interpret the results; this work is performed by experienced data mining experts. However, experts can develop strong decision-making instruments with appropriate artificial intelligence technologies, which can be used in the management of marketing. These instruments can help managers adopt proper strategies for marketing and manage campaign processes by making optimal decisions on time.

Recently, the rate of affirmative answers due to mass marketing campaigns has been very low (approximately 1%) due to global competition. Direct marketing campaigns focus on customers' willingness to purchase particular products or services [13]. However, the success of this type of campaign is limited. For example, in this data set, only 5,289 out of 45,211 customers gave positive responses during this direct campaign process. The success rate is approximately 11%, and there are some drawbacks to direct campaigns, such as those related to improving the negative attributes ascribed to banks by customers [10]. Thus, to overcome this problem, attractive, long-term deposit campaigns organized by banks should be managed more effectively. The main contributions of this work are:

1- Numerous data analysis studies on marketing can be found in the literature. However, most of these studies do not go beyond performing analyses. Even in the best cases, it is not possible for these analyses to be used as supportive instruments in decision making tasks whereby managers develop their marketing strategies. One cannot expect managers to conduct analyses, which involve advanced technologies, and interpret the results; this work is performed by experienced data mining experts. However, they can develop strong decision-making instruments with appropriate artificial intelligence technologies, which can be used in the management of marketing. For this reason, this study aims to develop an intelligent instrument to help managers adopt proper strategies for marketing and manage campaign processes by making optimal decisions on time.

2- This study focuses on producing solution to important problems dealing with direct bank marketing campaign such as very low success rates, risk of improving the negative attributes by customers to the campaigns. For this, DM approaches and expert system technologies are used together.

3- A large dataset (45211 records) was dataset from a Portuguese bank. The data, including the effects of the global financial crisis that peaked in 2008, were collected from 2008 to 2010.

4- Seven DM models (Neural Networks NN, Logistic Regression LR, Discriminant Analyses DA, Naive Bayes NB, K Nearest Neighbours KNN, Support Vector Machines SVM and Decision Trees DT) were used and compared their predictive performances with IBMMS. Also the phases of development intelligent system were shown using the DT model and its components.

5- The DT model, generated given best predictive performance with interpretable rules, was used to create inference engine of the system. The model reduced the original set to 9 attributes (out of 16).

6- IBMMS provides the manager guidance on how the campaign will be conducted by considering the situations of the customers. The manager can follow the customers as individuals or as a whole group on the decision tree and can make a campaign decision that leads to the desired response by the customers.

The study is organized as follows: bank marketing data presents in section 2. Information about the decision support systems in the field of marketing and their development to date is given in section 3. The phase of development IBMMS and information on its components is presented in section 4. Conclusions are drawn in section 5.

## 2. BANK MARKETING DATA

The UCI Machine Learning Repository provided the bank marketing data set for this study. The data were obtained from a Portuguese bank, which uses its own contact center to conduct direct marketing campaigns. Although the internet banking channel (e.g., by providing information to certain target clients) served as a secondary channel at times, human agents communicating over telephones was predominant. In addition, every campaign was managed in a unified manner, and the results for all channels were combined.

The dataset was obtained from 17 campaigns launched between May 2008 and November 2010. In the course of these phone campaigns, an appealing, long-term deposit application with good interest rates was proposed. The data relate to a Portuguese banking institution's direct marketing campaigns. These campaigns were conducted using phone calls. To determine if the product (bank term deposit) was subscribed to, there was frequently a need to contact the same client more than once. The goal of the classification was to estimate if the client subscribed to a term deposit (variable y). This data set consists of 45,211 records and 17 attributes (Table 1).

Table 1. Bank marketing data set descriptions

|  | Attributes | Description |
|---|---|---|
| **Bank client data** | | |
| 1 | Age (numeric) | |
| 2 | Job (categorical) | Job descriptions |
| 3 | Marital (categorical) | Married, divorced, etc. |
| 4 | Education (categorical) | Unknown, secondary, primary, tertiary |
| 5 | Default | Yes or no (has credit card?) |
| 6 | Balance (numeric) | Average yearly balance |
| 7 | Housing loan | Yes/no |
| 8 | Personal loan | Yes/no |
| **Related with the last contact of the current campaign** | | |
| 9 | Contact type (categorical) | Unknown, telephone, cellular |
| 10 | Day (numeric) | Last contact day of the month |
| 11 | Month (categorical) | Last contact month of year |
| 12 | Duration (numeric) | Last contact duration (second) |
| **Other attributes** | | |
| 13 | Campaign (numeric) | Number of contacts performed during this campaign and for this client (numeric, includes last contact) |
| 14 | Pdays (numeric) | Number of days that passed by after the client was last contacted from a previous campaign |
| 15 | Previous (numeric) | Number of contacts performed before this campaign |
| 16 | Poutcome (categorical) | Outcome of the previous marketing campaign |
| **Output variable (desired target)** | | |
| 17 | Y (binary) | Has the client subscribed a term deposit? (yes/ no) |

# 3. DECISION SUPPORT SYSTEMS IN MARKETING

Marketing decision support systems may be created to provide support for a variety of decision-making matters and problems at various levels [14]. It is clear that the essential goal of MDSS is to provide support for decision making in a strategic planning context, which comprises research and development, product design and planning, customer profile analysis, large-scope forecasting, etc. [15] (Rao, 2000). The "what-if" capability, which is common among the majority of DSSs, lets managers discover the consequences of a certain strategy or set of options prior to allocating time and money to the strategy [16]. These simulation models support what-if scenario analyses, which cast light on important business factors and variables that affect various marketing strategy options [17].

The role of DS methods and models for marketing decisions has been significant since decision DSSs were first developed [18]. Over the last five decades, different methods such as optimization, knowledge-based systems, and simulations have been developed. These methods have been found to be useful in many marketing domains such as new product development, pricing, and advertising [19]. Customer segmentation or profiling is accepted as an significant area among these marketing domains [20, 21, 22, 23]. There are two reasons behind this behaviour. The first reason is the customer-centric nature of the marketing paradigm [24] and of appropriately targeted marketing and services.

Second, unsolicited marketing is expensive and inefficient (e.g., low response rate) [25]. Additionally, increased effort has been focused on collecting and analysing customer data to help make better marketing decisions. The development of internet shopping technologies and database systems has sped up this approach. Data mining has been a significant instrument in this respect. Different DM methods, including statistical analyses and ML algorithms, can be instrumental in effective targeted marketing and customer segmentation [22]. Especially, classical DM algorithms can provide better prediction in terms of both accuracy and cumulative gains [26]. Various advanced technologies have become available to provide better MDSS for spec situations.

An intelligent system consists of a knowledge base, an inference engine, and a user interface. In the knowledge base, facts and knowledge about a firm's goals and strategies are catalogued. The knowledge is represented by a rule-based scheme in the form of a series of "if-then" condition-action statements. The knowledge is used by the inference engine, which processes the knowledge to draw conclusions.

# 4. DEVELOPMENT OF INTELLIGENT BANK MARKET MANAGEMENT SYSTEM

Detailed research into market management was performed prior to the development of IBMMS. This research indicated that many studies have been conducted on marketing, but most of them could not go beyond data analyses. Additionally, various analytical studies, such as studies on expert systems, have been conducted in this field, but their number remains insufficient. Nevertheless, an expert system should have a knowledge base, inference engine and user interface. Thus, IBMMS was developed as an supportive tool to be able to use in decision making while managers develop their marketing strategies.

Various expert systems and decision support systems have been developed in the field of marketing, but their inference engines generally consist of rules based on expert knowledge. Thus, they cannot surpass the performance of experts. However, it is possible to combine the successes of data mining and expert systems. Therefore, the systems to be developed in this way will be more powerful than other systems. However, unfortunately, this system has not yet been developed for the management of marketing campaigns.

For this reason, this study is devoted to developing an intelligent system with capabilities of interpretation, guidance and decision making in the marketing field. This section describes how to develop IBMMS system benefiting from data mining analyses.

The Intelligent Bank Market Management System (IBMMS) was developed to increase the efficiency of telemarketing campaigns by determining the main characteristics that affect success and, thereby, contribute to the better management of available resources (e.g., phone calls, human effort, and time) for obtaining a high-quality and affordable set of potential customers. There are five components (Fig. 1) that comprise IBMMS: the analysis, knowledge base, inference engine, advisor, and user interface.

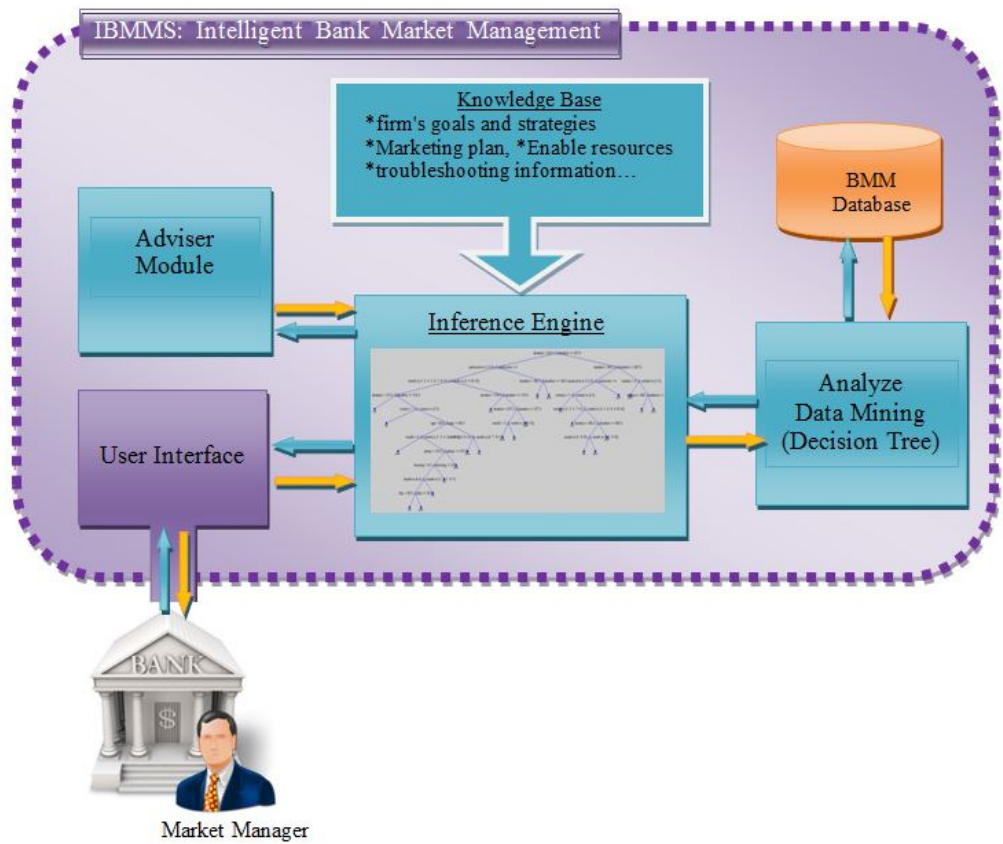

Figure. 1. IBMMS system' structure and components

## 4.1. Decision Tree Learning Algorithm for Knowledge Discovery

Different data mining methods (Neural Networks, Logistic Regression, Discriminant Analyses, Naive Bayes, Support Vector Machines and Decision Trees) are applied to the bank marketing data set, and their classification performances are compared in part 4.7. According to the results obtained from the analyses of the bank market dataset, the decision tree method is found to be more appropriate than the others for the development of an intelligent system.

*Knowledge discovery* is the name of the general process of obtaining useful knowledge from the data. In contrast, *data mining* represents a particular step in this process.

Acquiring rules in the form of "if-then" decision tree schemas is possible using this algorithm. The most important advantage of decision tree methods is based on compactness and clearness of presented knowledge and high accuracy of classification.

We can define a decision tree as a basic structure that performs classification tasks. Inside a decision tree, each node represents an attribute X, and each branch stands for this attribute. The anticipated class variable is Y in this study (Table 2).

$X_i$ represents each attribute: age, job, marital status, education, default, balance, housing, loan, contact, day, month, duration, campaign, pdays, previous, and poutcome. Y represents if the client subscribed to a term deposit (Yes/No).

$(x,Y) = (x_1, x_2, x_3, x_4, x_5, x_6, x_7, x_8, x_9, x_{10}, x_{11}, x_{12}, x_{13}, x_{14}, x_{15}, x_{16}, Y)$

Table 2. Example records (x,Y) from database

| $x_1$ | $x_2$ | $x_3$ | $x_4$ | $x_5$ | $x_6$ | $x_7$ | $x_8$ | $x_9$ | $x_{10}$ | $x_{11}$ | $x_{12}$ | $x_{13}$ | $x_{14}$ | $x_{15}$ | $x_{16}$ | Y |
|---|---|---|---|---|---|---|---|---|---|---|---|---|---|---|---|---|
| 22 | 12 | 2 | 2 | 0 | 10971 | 0 | 0 | 2 | 28 | 6 | 181 | 2 | -1 | 0 | 1 | no |
| 57 | 5 | 1 | 4 | 0 | 1884 | 0 | 0 | 3 | 28 | 6 | 133 | 4 | -1 | 0 | 1 | no |
| 60 | 6 | 1 | 2 | 0 | 2060 | 0 | 0 | 2 | 29 | 6 | 135 | 4 | 95 | 4 | 2 | no |

Generally, the decision tree algorithm consists of 5 steps.

*Step 1*. The training data are represented as "S". If the attributes are continuously valued, they should be categorized. These categories may be called with identifier labels. Later, the training set (S) is placed in a single tree node.

*Step 2*. If all patterns in S belong to the same class, stop.

*Step 3*. Select the attribute that gives the "best" split. The data set is split into smaller subsets using specific split criteria. Each of these subsets is split recursively until all of them become pure (when the cases in each subset belong to the same class) or until it is not possible to further their "purity".

*Step 4*. Split the node according to the values of the class variable. The tree's terminal nodes are created in this way. The splitting criteria cause the major difference between the DT construction procedures. The Gini Index is generally used as the splitting criteria.

*Step 5*. When the following conditions are met, stop. Otherwise, go to step 3.
(a) If there are no other nodes to split or if all patterns of the subsets belong to a single class (b), there are no surviving attributes.

Some specific definitions and equations for the Decision Tree are given below.

The DT splits nodes based on either impurities or node errors. Impurities can have one of several meanings depending on one's choice of the Split Criterion name-value pair: Gini's Diversity Index (gdi). The node's Gini Index is $\sum_i p^2(i)$ [27].

The sum is over the classes i of the node and p(i) is the observed fraction of classes of class i that reach the node. A node with a single class is pure node and its Gini Index value is "0", otherwise, positive. So, this index is a node impurity's measure.

The Gini index with p(i) can be described as a node's deviance in bellow.

$$- \sum_i p(i) \log p(i)$$

For a pure node is a deviance "0", otherwise, positive.

Another measure used to split node is Twoing rule [28]. The fraction of members of class i in the left child node after a split is represented with L(i), and the other side (right) with R(i). It is chosen to maximize split criterion.

$$P(L)P(R)\left(\sum_i |L(i) - R(i)|\right)^2$$

P(L) and P(R) are the observations' fractions splitting to the left and right. If the statement is large, the splitting made each child node more pure. If the statement is small, the splitting made each child node similar to each other nodes and, and so similar to the parent node.

The node error is described as the misclassified classes' fraction at a node. The node error is 1-p(j) for the class with the largest number of training patterns at a node.

Matlab software was used for implement the DT method in this study. Data analysis was made by using following code. This analysis study was showed that nine attribute (duration, poutcome, month, balance, contact, age, pdays, day, housing) are more important than others in decision making. Thus, they placed on the DT schema and used to develop inference engine of IBMMS.

```
MATLAB software code used to implement the DT
>>Ver=importdata('bankaveri.xlsx);
>>vars={'age' 'job' 'marital' 'education'  'default' 'balance' 'housing' 'loan' 'contact' 'day' 'month'
       'duration' 'campaign' 'pdays' 'previous' 'poutcome'};
>>x= [age job marital education default balance housing loan contact day month duration
campaign pdays previous poutcome];
>>y=strcat(clsout,{});
>>t=classregtree(x,y,'method', 'classification', 'names', vars,'categorical', [2 3 4 9 11 16], 'prune',
'off', 'crossval', 'on');
>>yPredicted=eval(t,x);
>>cm=confusionmat(y,yPredicted);
>>N=sum(cm(:));
>>err=(N-sum(diag(cm)))/N;
>>tt=prune(t,'level',2);
>>tt=prune(t,'level',100);
```

The DT classifier created via MATLAB tool generates "no" result according to the pattern vector (16 input) in bellow.

```
>> orn=[ 41 7 3 2 0 270 1 0 1 5 5 222 1 -1 0 1];
>>guest =eval(tt,orn)
guest =    'no'
```

## 4.2.Analyse Component

The bank marketing data sets are raw, unprocessed data sets. The data mining procedure requires the coding of information belonging to the customers. For example, the job attribute is composed of categorical information. Admin, unknown, unemployed, management, housemaid, entrepreneur, student, blue-collar, self-employed, retired, technician, and services are coded with 1, 2, 3, 4, 5, 6, 7, 8, 9, 10, 11, and 12, respectively, and digitized for the analysis. The transformed data are now ready for data mining. In this study, we benefited from rules of DT classifier created by using Matlab tool to design analysis component of IBMMS. A few decision tree models obtained as a result are processed by specialists during interpretation evaluation, producing the IBMMS inference engine. The complete process is schematized in Fig. 2.

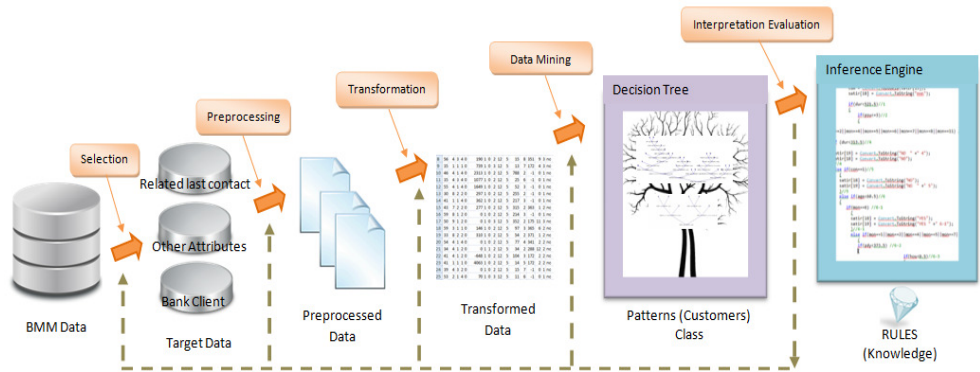

Figure 2. Knowledge discovery for inference engine

The inference engine obtained using an analysis module helps managers make decisions on various matters, such as successfully managing the campaign, determining customer trends, and specifying the time to contact customers and launch the campaigns. The analysis module facilitates the availability of a customer portfolio list, which follows paths on the decision tree as an advisor component. This module is an important information source for the decisions being offered by the advisor component.

## 4.3. Knowledge Base component

A knowledge base is an information repository that allows for the collection, organization, sharing, searching and use of information. The knowledge base can be machine-readable or intended for human use. Human-readable knowledge bases allow users to retrieve and utilize the contained knowledge. Common uses of these knowledge bases include complementing a help desk or sharing information among workers within an organization.

A knowledge base component has been created within IBMMS whereby a substantial amount of information, such as campaign plans, strategies, and documents related to the campaign, are shared and kept. IBMMS reports and customer lists can be kept during and at the end of the campaign. Thanks to the knowledge base, managers can have control over the process during each step of campaign, and thus, they can make correct decisions.

## 4.4. Inference Engine Component

IBMMS includes an intelligent component with an inference engine. This engine contains rules (if ..then ..) derived from schemas of the Decision Tree algorithm in the analysis component. These rules help the system provide information during a campaign on customer preference, following strategy, etc.

Although decision trees exhibit a better performance compared to many DM algorithms, the decision schemas cannot be directly used in the inference engine. To create the best decision rules, an expert must interpret the decision trees. It may even be necessary to combine a few decision tree models for the inference engine to maximize the classification performance. Sample codes from inference engine are given below.

```
else if(dur < 605.5) //1 - 3
 {
        if(mon == 5||mon == 8||mon == 10)
          {
             satir[18] = Convert.ToString("NO");
             satir[19] = Convert.ToString("NO" + "1 - 5");
          }
```

```
else if(mon == 3||mon == 6||mon == 9 || mon == 12)
      {
        satir[18] = Convert.ToString("YES");
        satir[19] = Convert.ToString("YES"   + "1 - 6");
      }
else { satir[19] = Convert.ToString("undefined"); };

        }
    else
      {
      satir[18] = Convert.ToString("YES");
      satir[19] = Convert.ToString("YES" + "1 - 3");
      }
 }
```
……….

## 4.5 Advisor Component

This component provides guidance on how the campaign will be conducted by considering the situations of the customers. The manager, using an advisor, can follow the customers as individuals or as a group on the decision tree and can make a campaign decision that leads to the desired response by the customers. During the campaign, it is possible to intervene in certain aspects, whereas this is not possible for others. To illustrate, the age of the customer is a demographical feature that cannot be modified, but implementing the campaign in the 6[th] month is a positive strategy in terms of the success of the campaign for the group of customers under 60 years of age according to the advisor.

Thanks to the advisor, the manager can see suggested strategic paths at each node of the decision tree and can specify required strategic decisions for the successful execution of the campaign. Thus, each node is a decision threshold. The strategy that the manager will specify at this point (Fig. 3) will alter whether the customers being targeted by the campaign will respond "YES" or "NO".

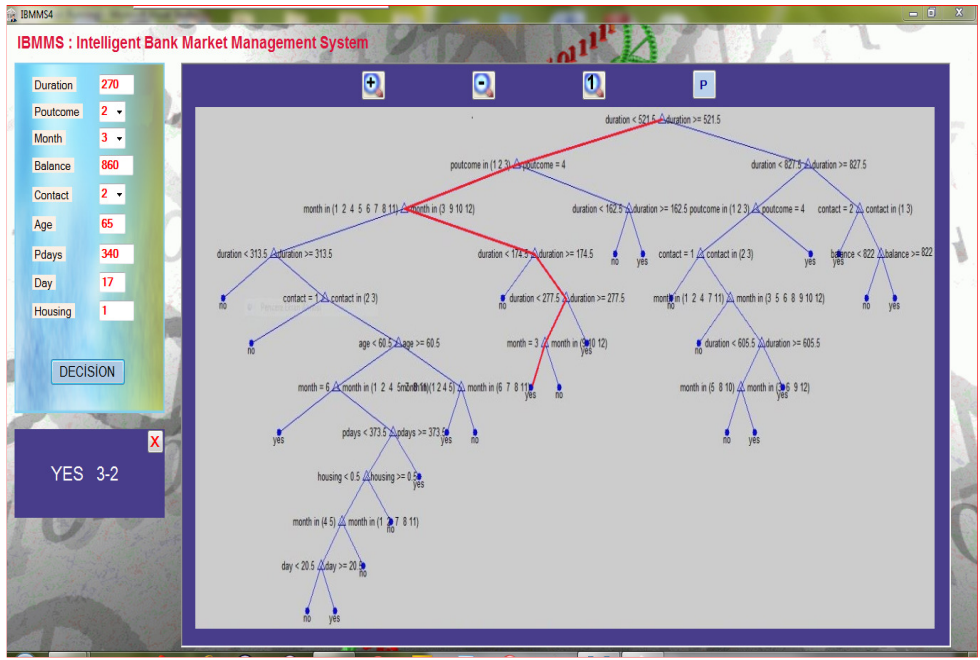

Fig. 3. Strategic path showed by IBMMS

At certain nodes, there is no need to see more detailed information about the customer. For example, there is no need for certain information, such as the month when the campaign will be launched, contact status, age and housing, from customers who gave negative responses to previous campaigns. In this case, one needs to determine the correct contact duration for contacting these customers. The advisor guides the manager on the necessary actions (Fig. 4).

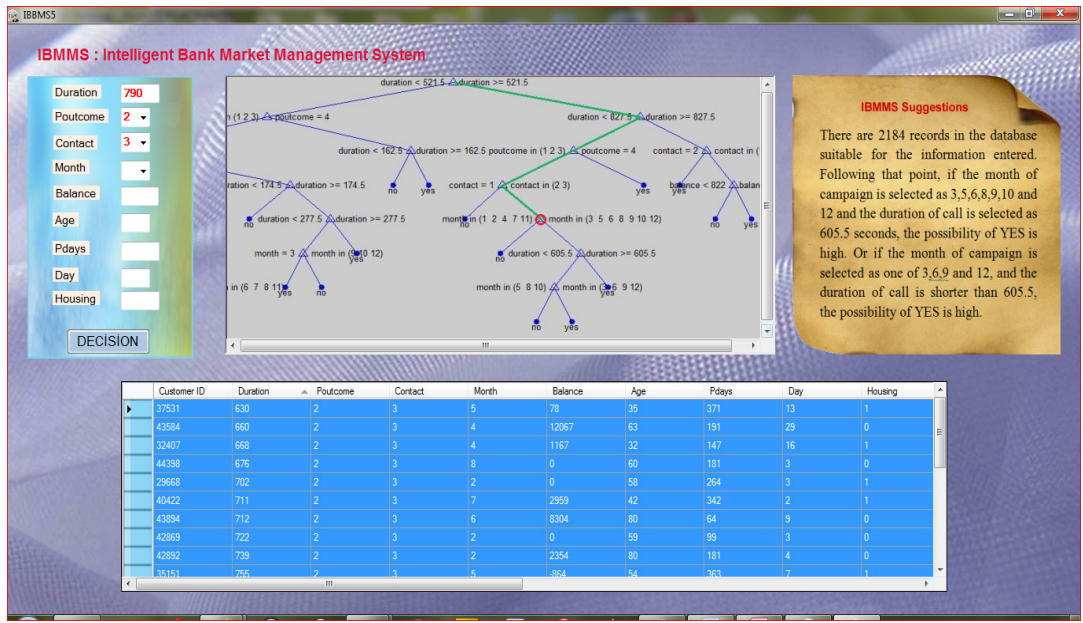

**Figure 4.** A decision threshold and advisor's suggestion to the manager

The advisor shows the customer information to the manager during and at the beginning of the campaign and marks the customer's position on the decision tree. The advisor helps the manager specify the right strategy and make the right decision using the roadmap obtained from the decision tree for the customer/customers. The manager can also create imaginary customers if he feels that this is necessary. He can forecast the potential results by modifying these data (Fig. 3). The advisor can identify and list all of the customers suitable for the case using only three pieces of information, although there are normally 16 features for each customer. Moreover, the advisor indicates the decision path suitable for this situation by drawing a green line over it. Following the identified decision threshold, the advisor can provide the manager with a suggestion *"IBMMS suggestion: There are 2184 records in the database suitable for the information entered. Following that point, if the month of campaign is selected as one of 3,5,6,8,9,10,12 and the duration of call is selected as 605.5 second, the possibility of YES is high. Or if the month of campaign is selected as one of 3,6,9,12, and the duration of call is shorter than 605.5, the possibility of YES is high"* that results in positive responses to the campaign from the customer (Fig. 4). In addition, the advisor can directly provide the result when nine variables are entered, and the advisor marks the suitable path for the decision by marking it in red up to the decision (Fig. 3).

## 4.6. User Interface

IBMMS has a user-friendly interface. The results and situations obtained from the communication and interaction of all components are presented to the customer via this interface. The interaction among the components of the system is provided via this interface. The entire system is Windows-based and has been developed using Visual Studio. NET, ASP, and C#. The database operations were performed using SQL server 2012.

## 4.7. Comparison of IBMMS and Other DM Models

In decision support systems, reliability is very important. Two measures are used to separately evaluate a classifier's performance on different classes: sensitivity and specificity (Table 3). In the following section, we calculate the IBMMS inference engine performance using 45,211 customers.

Table 3. IBMMS classification performance using the current data

| IBMMS | Predicted class | | | | | |
|---|---|---|---|---|---|---|
| | No | | Yes | | Sum | |
| NO | TN=37735 | 83.46% | FN=697 | 1.54% | 38432 | 85.00% |
| YES | FP=2187 | 4.84% | TP=4592 | 10.16% | 6779 | 15.00% |
| Sum | 39922 | 88.30% | 5289 | 11.70% | 45211 | 100.00% |

$$Sensitivity = \frac{True\ Positive\ (TP)}{TP + False\ Negative\ (FN)} = \frac{4592}{4592 + 697} = 0.87$$

This sensitivity rate shows the effect of previously determining customers that will respond positively to the campaign.

$$Specificity = \frac{True\ Negative\ (TN)}{TN + False\ Positive\ (FP)} = \frac{37735}{37735 + 2187} = 0.95$$

In contrast to the sensitivity, the specificity rate shows the effect of previously determining customers that will respond negatively to the campaign. However, these rates are the results computed using the IBMMS inference engine using the current IBMMS data set. However, market managers can change the customer's response using IBMMS and by considering its suggested strategies. Thus, managers can increase the success of the campaign and better direct this process.

In contrast to the sensitivity, the specificity rate shows the effect of previously determining customers that will respond negatively to the campaign. However, these rates are the results computed using the IBMMS inference engine using the current IBMMS data set. However, market managers can change the customer's response using IBMMS and by considering its suggested strategies. Thus, managers can increase the success of the campaign and better direct this process.

A Different data mining methods were also applied  (Neural Networks NN, Logistic Regression LR, Discriminant Analyses DA, K Nearest Neighbours KNN, Naive Bayes NB, Support Vector Machines SVM and Decision Trees) to the bank marketing data with the aim of providing a comparison in terms of the predict performance. These methods often fit the data well, but because of a lack of comprehensibility, they are considered to be black box technologies. The data set is divided into a training and test set. To calibrate the classifier parameters, the training set was used. The trained classifier was used to make predictions for the test set. Cross validation is generally used to compare the performance of different classifier models. Here, holdout validation has been used; 40% of the data, selected randomly, is held for the test phase.

The predict performances of these classifiers were calculated and compared, as shown in Table 4. For these data, although NN model yields a little better successful predicts than others it cannot be used to develop a system because of its "black box" structure. Second best model DT is suitable for the development of an intelligent decision support system with comprehensible rules.

Table 4. Comparison of DM models' predict performances with IBMMS (bold denotes best value)

| Client Resp. | Predict (%) | Data Mining Methods | | | | | | | IBMMS |
|---|---|---|---|---|---|---|---|---|---|
| | | NN | LR | DA | KNN | NB | SVM | DT | |
| NO | Success | **98.32** | 97.53 | 90.25 | 93.87 | 92.02 | 91.56 | 94.07 | **98.19** |
| | Failure | 1.68 | 2.47 | 9.75 | 6.14 | 7.98 | 8.45 | 5.93 | 1.81 |
| YES | Success | **78.29** | 64.83 | 51.80 | 62.13 | 46.26 | 52.70 | 51.61 | 67.74 |
| | Failure | 21.71 | 35.17 | 48.20 | 37.87 | 53.74 | 47.29 | 48.39 | 32.26 |
| Accuracy (%) | | **88.31** | 81.18 | 71.03 | 77.99 | 69.14 | 72.13 | 72.84 | **82.96** |

## 5. CONCLUSION

Data mining studies attempting to solve many problems in the banking field can be found in the literature. However, these studies could not go beyond performing analyses. Only experts in the field can develop AI data mining techniques and determine how to put them into practice, how to interpret them and how to use them for decision making. In this study, the expert must have knowledge of both analysis techniques and implementation in the field; however, possessing both of these traits is extremely difficult. A marketing manager of a bank cannot be expected to have knowledge about these matters. Moreover, most analysis techniques do not yield the best results in terms of the initial classification or forecasting operation. The parameters of the analysis method evolve until the best model is obtained. These operations are mostly performed by the training and testing processes. The best model that yields the best solution to the problem at the end of the analysis cannot be presented to the managers or decision makers in the form that it is found. The best way to make this presentation possible is to transform the best models that can solve the given problems into systems that can be used by the decision makers. Until this transformation is realized, new technologies and models that are currently being studied can never be implemented in practice.

This study focuses on how all of the above process can provide usable information for managers or decision makers. IBMMS was developed for this purpose. The system shows a manager how he should manage a campaign according to the situation and according to what preferences customers will have and how the status of the customer/customers will evolve over the course of the campaign. The system provides assistance in the campaign management and decision-making process.

In the development of decision support systems, strong machine learning algorithms and expert system approaches should be used. Therefore, these decision support systems must be able to be converted into intelligent systems that can be used by decision makers as strong instruments in the decision-making process and for use when solving a variety of problems.

# REFERENCES

[1] Bhasin, M.L. (2007) Data Mining: a competitive tool in the banking and retail industries central council library. The Chartered Accountant, 55, 588-594.

[2] Putten, P.W.H. (2006) On data mining in context: cases, fusion and evaluation Proefschrift Amsterdam, Netherlands, Universiteit Leiden ISBN: 978-90-8891143-9, 2010.

[3] Kotler, P.G.A. (2006) Principles of Marketing. Pearson Prentice Hall.

[4] Bult, J.R. and T. Wansbeek (1995) Optimal selection for direct mail, Marketing Science, 14, 378–394.

[5] Gonul, F. and M.Z. Shi (1998) Optimal mailing of catalogs: a new methodology using estimable structural dynamic programming models, Management Science, 44: 1249– 1262.

[6] Piersma, N. and J. Jonker (2000) Determining the direct mailing frequency with dynamic stochastic programming. Tech. Rep. EI2000-34A, Econometric Institute, Erasmus University, Rotterdam, Netherlands.

[7] Kim, Y., W.N. Street, G.J. Russell and F. Menczer (2005) Customer targeting: a neural network approach guided by genetic algorithms, Management Science, 51, 264-276.

[8] Domingos, P. and M. Richardson (2001) Mining the network value of customers, In: Proc. of 7th ACM SIGKDD Int'l Conf, On Knowledge Discovery and Data Mining (KDD-01), ACM, New York, 57– 66.

[9] Nie, G., W. Rowe, L. Zhang, Y. Tian and Y. Shi (2011) Credit card churn forecasting by logistic regression and decision tree, Expert Systems with Applications, 38, 15273–15285.

[10] Moro, S., R. Laureano and P. Cortezp (2011) Using Data Mining for Bank Direct Marketing: an application of the CRISP-DM methodology, In P. Novais et al. editors. Proceedings of the European Simulation and Modelling Conference – ESM, Guimarães, Portugal, EUROSIS 2011, 117-121.

[11] Tsai, C.F. and Y.J. Chiou (2009) Earnings management prediction: a pilot study of combining neural networks and decision trees. Expert Systems with Applications, 36, 7183–7191.

[12] Mitchell, T. (1997) Machine learning. New York: McGraw-Hill.

[13] OU, C., C. LIU, J. HUANG and N. ZHONG (2003) On Data Mining for Direct Marketing, Proceedings of the 9th RSFDGrC conference 2003, 491–498.

[14] Cassie, C. (1997) Marketing decision support systems, Industrial Management & Data Systems, 97, 293-297.

[15] Rao, S.K. (2000) Marketing decision support systems for strategy building. Marketing Health Services, 20, 14-8.

[16] Sisodia, R.S. (1992) Marketing information and decision support systems for services. Journal of Services Marketing, 6, 51-64.

[17] LI, S., and B.J. Davies (2001) Key issues in using information systems for strategic marketing decisions, Int. J. Management and Decision Making, 2, 16-34.

[18] Mcdaniel, R. (2007) Management strategies for complex adaptive systems, Performance Improvement Quarterly, 20, 21–42.

[19] Hart, M. (2008) Systems for supporting marketing decisions, In: Burstein F, Holsapple C, editors. Handbook on Decision Support Systems, 395–418.

[20] Jonker, J., N. Piersma and R. Potharst (2006) A decision support system for direct mailing decisions, Decision Support Systems, 42, 915–925.

[21] Kim, Y. and W.N. Street (2004) An intelligent system for customer targeting: a data mining approach, Decision Support Systems, 37, 215–228.

[22] Mckelvey, B. (1999) Avoiding complexity catastrophe in coevolutionary pockets: strategies for rugged landscape, Organization Science, 10, 294–321.

[23] Verhoef, P., P. Spring, J. Hoekstra and P. LEEFLANG (2002) The commercial use of segmentation and predictive modeling techniques for database marketing in the Netherlands. Decision Support Systems, 34, 471–481.

[24] Vargo, S. and R. Lusch (2004) Evolving to a new dominant logic for marketing. Journal of Marketing, 68, 1–17.

[25] Gopal, R. (2001) Ad mediation: new horizons in effective email advertising, Communications of the ACM, 19, 17–30.

[26] Olson, D. L. and C. Bongsug(Kevin) (2012) Direct marketing decision support through predictive customer response modeling, Decision Support Systems,54,443-451.

[27] Coppersmith, D., S.J. HONG, and J.R.M. HOSKING (1999) Partitioning Nominal Attributes in Decision Trees, Data Mining and Knowledge Discovery, 3, 197–217.

[28] Breiman, L., J. Friedman, R. Olshen, and C. Stone (1984) Classification and Regression Trees. Boca Raton, FL: CRC Press.